# Using Vision Language Models for Safety Hazard Identification in Construction

Muhammad Adil, Gaang Lee, Vicente A. Gonzalez, Qipei Mei*

Department of Civil and Environmental Engineering, University of Alberta, Edmonton T6G 2R3, Alberta, Canada

*Corresponding author's email address: qipei.mei@ualberta.ca

**Abstract**

Safety hazard identification and prevention are the key elements of proactive safety management. Previous research has extensively explored the applications of computer vision to automatically identify hazards from image clips collected from construction sites. However, these methods struggle to identify context-specific hazards, as they focus on detecting predefined individual entities without understanding their spatial relationships and interactions. Furthermore, their limited adaptability to varying construction site guidelines and conditions hinders their generalization across different projects. These limitations reduce their ability to assess hazards in complex construction environments and adaptability to unseen risks, leading to potential safety gaps. To address these challenges, we proposed and experimentally validated a Vision Language Model (VLM)-based framework for the identification of construction hazards. The framework incorporates a prompt engineering module that structures safety guidelines into contextual queries, allowing VLM to process visual information and generate hazard assessments aligned with the regulation guide. Within this framework, we evaluated state-of-the-art VLMs, including GPT-4o, Gemini, Llama 3.2, and InternVL2, using a custom dataset of 1100 construction site images. Experimental results show that GPT-4o and Gemini 1.5 Pro outperformed alternatives and displayed promising BERTScore of 0.906 and 0.888 respectively, highlighting their ability to identify both general and context-specific hazards. However, processing times remain a significant challenge, impacting real-time feasibility. These findings offer insights into the practical deployment of VLMs for construction site hazard detection, thereby contributing to the enhancement of proactive safety management.

## 1. Introduction

The construction industry has a long history of being one of the most hazardous industries, indicated by the highest rates of accidents and fatalities recorded globally [1]. In the United States, the construction sector experienced 1,075 fatalities in 2023, the highest among all industries [2]. This corresponds to a fatal injury rate of 12.9 per 100,000 full-time equivalent (FTE) workers, significantly higher than the national average of 3.2 [2]. Similarly, in Canada, fatal accidents in the construction industry accounted for nearly 18.5% of all reported worker fatalities in 2022 [3]. These figures, often resulting from inadequate safety management and ineffective hazard identification, underscore the urgent need to address challenges such as the limitations of traditional manual monitoring and more recent computer vision-based monitoring approaches.

The dynamic and complex nature of construction sites contributes to inherent safety risks [4]. Hazards at the construction site exhibit spatiotemporal dynamics due to the continuous movement of machinery and workers, as well as the evolving nature of tasks throughout different construction phases [5]. As the project progresses, the quantity and types of hazards—such as large excavators or foundation pit edges—can decrease or disappear, and the locations of these hazards, like floor edges and openings, shift with the changing levels of the building structure. These dynamic and continuous variations in risk factors require systematic hazard detection capable of adapting to dynamic and varying site conditions.

Traditional safety personnel-dependent approaches for site safety monitoring fall short due to their manual nature and are expensive, time-consuming, and lack continuous monitoring. Computer vision methods have automated hazard detection, significantly improving efficiency and accuracy. Object detection, a vision-based technique for localizing and classifying predefined objects [6], has shown substantial progress in accuracy and processing speed, with numerous studies [7–10]. Naturally, there have been many studies applying object detection to detect safety hazards from images of construction sites by recognizing hazard-related entities, such as workers, their personal protective equipment (PPE), heavy equipment, and explosives [11,12]. However, these previous efforts are significantly limited in comprehensively identifying different types of hazards in construction scenes as construction hazard detection requires not only detecting hazard-related entities, but also incorporating the dynamic contexts and semantics of the detected entities [13], such as activity monitoring (e.g., detecting unsafe ladder use), behavior recognition (e.g., identifying signs of worker fatigue, unsafe posture) and interaction analysis (e.g., evaluating worker proximity to heavy machinery, potential struck-by hazards, or unstable structures). The trace intersecting theory suggests that accidents occur when both unsafe human acts and unsafe environments or conditions coincide within the same time and space[14,15]. Therefore, effective hazard identification requires not only detecting construction entities but also reasoning about the construction scene by analyzing the semantic relationship among various elements. This understanding of semantic relationships among different elements can help transform unstructured visual data into structured textual descriptions, enabling integration with external knowledge (e.g., safety regulations) for hazard identification [15]. More recently, researchers have explored image and dense captioning, computer vision techniques that generate textual descriptions for entire images or multiple regions within them. These methods have been integrated with Natural Language Processing (NLP) techniques to generate semantically rich descriptions for worker activities, interactions, and heavy machinery working scenarios and assess compliance with safety regulations [8,9,13]. However, despite improved semantic richness, dense captioning techniques still face challenges in scalability and adaptability similar to object detection. This is primarily due to their continued reliance on large-scale manually annotated, site-specific datasets, limiting its broader application in real-world construction hazard detection scenarios.

Recent advancements in conversational and generative artificial intelligence (AI) e.g., ChatGPT have found significant applications in the Architecture, Engineering, and Construction (AEC) industry [16,17]. Vision Language Models (VLMs) represent a powerful class of multi-modal AI systems capable of simultaneously processing both visual and textual information [18]. VLMs combine the visual understanding capabilities of computer vision models with the contextual reasoning and knowledge representation power of large-scale language models. This integration allows them to generate descriptive outputs, reason about object relationships, and interpret complex scenes based on both image content and textual knowledge.

To improve construction safety monitoring through comprehensive construction scene analysis, we propose and validate a VLM-based framework for construction hazard detection and risk assessment on construction sites. The framework begins by processing safety guidelines through a dedicated prompt engineering module, where the safety guideline text is analyzed and transformed into a structured prompt using a VLM. This prompt is designed to guide the VLM in hazard identification and proposing recommendations for mitigation. Once the prompt is generated, it is passed to VLM along with the construction site image. The VLM processes the visual information in the image according to the regulatory guidelines, generating descriptive outputs, which include identified hazards and their severity level, and providing recommendations for mitigation. This synergistic framework automates safety inspections, offering a more comprehensive and adaptive hazard identification process that potentially improves workplace safety.

## 2. Related Works

### 2.1. Computer vision in construction safety management

Construction sites are inherently complex and dynamic, characterized by numerous hazards that evolve as the project progresses [19]. Hazards on these sites are primarily caused by two key factors: workers' unsafe behaviors [20] and the constantly changing conditions and activities at the project sites, which often result in overlooked safety hazards that lead to preventable occupational injuries [21]. Manual methods for safety inspection and management can be costly, labor-intensive, and lack continuous monitoring [7]. Additionally, documenting any potentially hazardous actions or events adds to managerial and administrative workload, further straining resources and reducing efficiency [22].

In recent years, deep learning techniques have been deployed widely on construction projects and have shown significant capabilities in the identification of workplace hazards [7]. Entity detection involves detecting and localizing key elements within an image and has been the focus of numerous studies in workplace hazard detection over the past decade [9]. These studies have primarily targeted safety-related elements such as workers, PPE, tools, onsite objects and machinery, etc. Algorithms like You Only Look Once (YOLO) [23], faster region-based CNN (Faster R-CNN) [24], and mask region-based CNN (Mask R-CNN) [25] have been extensively evaluated for this purpose. The shift toward leveraging advanced deep learning frameworks for construction safety management has enhanced the ability to detect and interpret hazards in dynamic construction environments and has offered a more efficient and accurate alternative to traditional methods [7]. For example., Nath et al. [10] adopted the third version of YOLO to verify PPE compliance by detecting the helmets and vests of workers in real-time. Lee et al. [26] utilized the fifth version of YOLO to detect small tools in indoor construction sites. Chian et al. [27] leveraged a missing object detection approach to automatically detect missing barricades on construction sites. While these methods have achieved substantial progress in object detection and localization, they remain limited to recognizing and tracking individual entities without interpreting their contextual relationships [28]. Such limitations hinder their ability to capture high-level semantics, such as the interactions between workers and objects.

Semantic richness is essential for comprehending the complex dynamics of a construction site scene, enabling a deeper understanding of interactions between workers and objects. This level of contextual detail is important for identifying safety risks linked with improper actions and analyzing complex spatial-temporal relationships [8,29]. To address this gap, recent studies have explored image captioning and dense captioning [30] techniques that provide semantically rich descriptions for entire images or specific regions within them. For example, Zhang et al. [15] developed an image caption module by integrating interaction labels into a deep-learning network for object detection to generate a scene graph. Similarly, Wang et al. [13] integrated an object detection encoder and an image captioning decoder to extract semantic information. While this approach enabled the representation of relationships between objects in a structured format, the interaction labels relied on predefined keywords rather than using descriptive natural language. Wang et al. [8] explored dense captioning to provide semantically rich captions for specific regions in construction images, enabling better analysis of worker behaviors and interactions. Despite their potential, dense captioning methods still face challenges in scalability and adaptability, as these methods often depend on large-scale manually annotated datasets. These datasets are resource-intensive to create and difficult to generalize across diverse construction scenarios, limiting the broader applicability in real-world construction hazard detection. To address this, there is a need for a more comprehensive approach that not only detects construction hazards but also their contextual implications.

*2.2. Natural Language Processing and its integration with computer vision for construction site safety*

The integration of natural language processing (NLP) with computer vision offers a promising direction, allowing for deeper reasoning by linking visual information with regulatory guidelines, incident reports, and domain knowledge. In the construction industry, data related to safety, such as regulations, guidelines, safety reports, and incident logs, is often stored in electronic text formats [8]. This information serves as a critical resource for identifying potential hazards and effective decision-making. NLP techniques can help analyze large amounts of textual data, such as regulations and safety incident reports, and extract valuable insights such as potential hazards and patterns [31]. By combining vision-based perception with language-based interpretation, NLP-enhanced frameworks can help automate compliance assessment in construction safety [8].

Researchers have explored various NLP techniques (e.g., document classification, knowledge extraction, and text mining) for construction and safety management [32]. Document Classification sorts documents into predefined labels. In safety management, document classification can categorize accident causes [33] and near-miss accidents [34] by analyzing incident reports. Knowledge extraction focuses on transforming complex, unstructured text into structured and actionable insights and extracting the information. For example, Wang and El-Gohary [35] deployed a CNN model to extract relevant content from safety documents and verify its compliance with the regulations. Text Mining, another NLP technique, leverages a statistical approach to understand patterns and correlations in textual data. It has been used in safety management studies to identify causes of fire incidents at construction sites by analysis of news articles [36] and to extract key risk factors from reports for safety accidents [37].

The integration of NLP with computer vision enhances site safety management by providing a comprehensive view of safety conditions, improving hazard detection, and supporting more effective risk mitigation strategies for managers [8]. Some studies have developed frameworks by linking detected objects with safety guidelines to automate safety inspections. For example, Zhang et al. [15] developed a hybrid framework for a combination of computer vision and domain knowledge for construction hazard detection. Their approach involved first detecting classes of objects from images and then utilizing a supervised BERT model and safety regulations to assess the hazard probabilities. Mei et al. [38] adopted a similar approach to feature extraction using deep learning models and then integrating them with knowledge graphs built upon an ontology model and safety regulations. Wang et al. [8] introduced dense captioning to extract meaningful textual descriptions of construction activities, which were then cross-referenced with safety regulations for automated safety assessment. Despite these advancements, existing NLP-integrated approaches still rely on predefined rule-based mappings or supervised learning models, limiting their ability to generalize across diverse construction scenarios.

*2.3. Limitations of existing methods*

While the aforementioned studies show the potential of computer vision and NLP technologies in construction safety management, these methods have several notable limitations:

(1) A key limitation of existing entity detection methods in computer vision is their inability to capture the semantic relationships within construction scenes. These methods focus on detecting individual objects and locations without considering how they interact or relate to each other in the broader context of the construction site [15,39]. This lack of contextual understanding can hinder the accurate assessment of the dynamic and complex situations commonly found in construction environments. For example, a tool detection model might identify a tool in a cabinet or ground, but it may fail to distinguish which of these situations poses a hazard.

(2) Another significant limitation is the heavy reliance on predefined classes and captions for objects and images [40]. Existing methods for construction hazard detection are models trained on specific categories and annotations, which restricts their ability to recognize hazards or scenarios that fall outside the scope of training data. Consequently, they are unable to detect new hazards or unforeseen risks that are not explicitly represented in their training datasets, reducing their adaptability to the evolving conditions of construction sites.

(3) The accuracy of these methods depends on the quality and diversity of the datasets used for training [40]. Large datasets are required for object detection, segmentation, image captioning, and dense captioning techniques frequently used in previous literature. However, creating such datasets is labor-intensive, time-consuming, and expensive [41]. This challenge, combined with varying site layouts, lighting conditions, and object appearances, makes it difficult to generalize the models effectively across different scenarios.

(4) Computer vision integrated with NLP has shown significant potential. For instance, the frameworks developed by Zhang et al. [15], Mei et al. [38], and Wang et al. [8], have demonstrated the ability to monitor safety compliance by referencing safety guidelines. However, these methods are semi-objective as they rely on extracting information from scenes and comparing this extracted information to safety guidelines based on semantic similarity, making them site-specific. This reliance on semantic matching means that if a construction site follows different safety guidelines, the computer vision model will require retraining with new annotated data. This challenge limits its generalizability and poses challenges for widespread adoption.

These limitations underscore the need for a more adaptable and context-aware approach to construction hazard detection—one that can dynamically interpret diverse construction environments without requiring extensive retraining.

*2.4. Leveraging VLMs for Construction Hazard Detection: Potential and Available Models*

Recent advancements in VLMs have shown the potential to detect construction hazards in a more adaptable and context-aware manner with large-scale language models' reasoning capabilities. Following the foundational success of LLMs such as GPT-3, and Llama 3, across diverse fields, including construction [16], the development of VLMs has emerged as a transformative advancement in artificial intelligence. VLMs combine the capabilities of visual encoders with the contextual understanding of LLMs, enabling models to process and reason over both visual and textual data [42]. Visual encoders in vision language models are typically trained on extensive image datasets using unsupervised methods to capture rich visual representations. These representations are then integrated into LLMs, which allows the models to generate outputs that combine visual perception with linguistic comprehension. Figure 1 shows an overview of a VLM architecture.

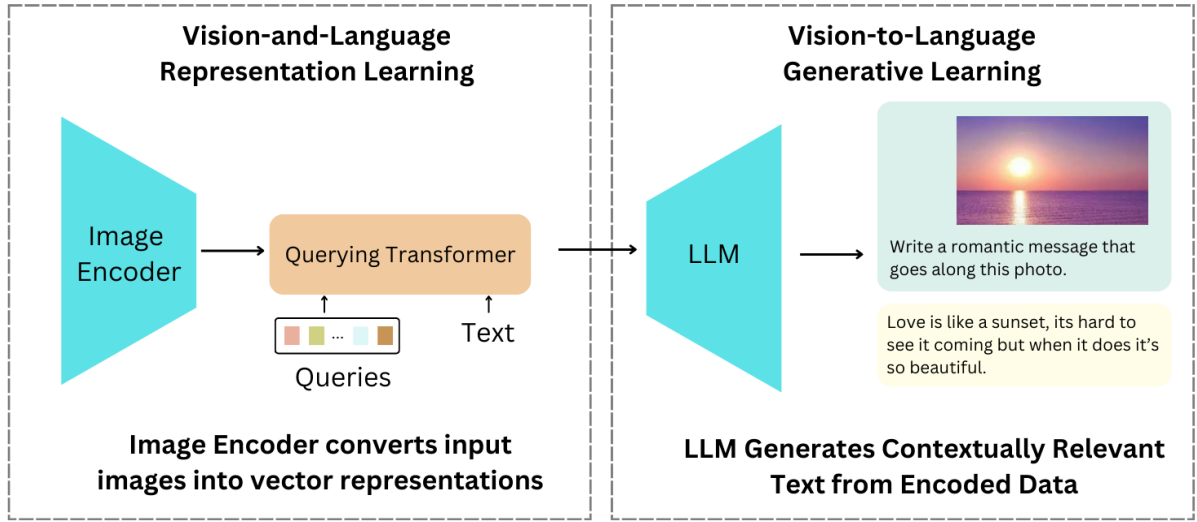

**Figure 1.** Overview of a VLM: BLIP-2 framework. Adapted from BLIP-2 Framework by Li et al. [43], licensed under CC BY 4.0.

By integrating vision and language reasoning, VLMs can go beyond object detection to understand spatial relationships, infer implicit hazards, and adapt to different site conditions without extensive retraining. Recent studies have explored the application of VLMs in the construction industry. Fan et al. [17] introduced ErgoChat, a visual-language model-based interactive system designed for ergonomic risk assessment of construction workers. Ersoz [44] explored the use of GPT-4V for construction progress monitoring and tracking development changes over time.

Recently, many powerful VLMs have been developed, each characterized by unique architectural innovations or training methodologies. In particular, these models have been made publicly available, allowing broader access and enabling their application across various domains. Below are the notable examples.

2.4.1. GPT-4o

OpenAI GPT models have consistently set new benchmarks in generative AI with novel approaches to process and generate text from visual and text data. GPT-4 is a multimodal model based on Transformer architecture and finetuned using Reinforcement Learning [45]. GPT-4o, its successor, builds upon the same foundational architecture with enhanced reasoning capabilities. OpenAI allows access to perform inference, fine-tune, and evaluate models using its API.

2.4.2. Gemini

Gemini models by Google DeepMind represent a significant advancement in the multimodal LLMs. Similar to other large language models, Gemini models leverage advanced optimization techniques to enhance scalability. Gemini 1.5 Pro, Gemini 1.5 flash, and the recently released Gemini 2 flash are also based on Transformer architecture and demonstrate exceptional performance in reasoning tasks, due to their higher context-length processing capabilities. Similar to GPT models, Gemini models can be accessed through Vertex AI API.

2.4.3. Llama 3.2

Meta AI's Llama 3.2 [46] extends the capabilities of the Llama model series by incorporating vision-based understanding into its framework. This multimodal model incorporates a Vision Transformer (ViT) as its vision encoder designed to extract detailed visual features, which are then aligned with its textual understanding using cross-modal attention mechanisms. Llama 3.2 Vision is available open-sourced in configurations with 11B and 90B parameters.

2.4.4. InternVL2

InternVL2 [47] series, developed by OpenGVLab, represents a significant advancement in open-source multimodal large language models. InternVL2 employs a ViT-MLP-LLM configuration, integrating ViT with Multilayer Perceptron (MLP) and LLMs to facilitate comprehensive visual-text processing [47]. The InternVL2 series offers models of varying scales, ranging from 1B to 108B parameters.

*2.5. Knowledge gaps and research objectives*

Despite these VLMs' availability and demonstrated success in different domains, to the authors' best knowledge, there is a notable lack of research investigating their superior capacity in construction hazard detection over previous computer vision- or NLP-based approaches, which stems from their adaptable and context-aware scene understanding. Specifically,

(1) It has not yet been examined whether VLMs are effective in identifying construction hazards. While VLMs have been extensively studied for general reasoning and multimodal tasks, their ability to detect and assess hazards in construction environments remains unverified.

(2) There is a lack of comprehensive evaluation comparing the accuracy, efficiency, and real-time feasibility of different state-of-the-art VLMs (i.e., GPT-4o, Gemini 1.5, Llama 3.2 Vision, and InternVL2) for construction hazard detection. Although these models have been benchmarked in various applications, their performance in terms of hazard identification accuracy, processing speed, and practical deployment in dynamic construction settings has not been systematically assessed.

To address these research gaps, this study aims to develop and validate an adaptable and context-aware-powered construction hazard detection framework utilizing state-of-the-art pre-trained VLMs. The framework analyzes the safety guidelines text and engineer prompts, detects potential hazards in construction images, checks regulatory compliance, and suggests risk mitigations across diverse construction scenarios. Specifically, the validation focuses on i) Assessing the feasibility of the proposed framework for construction hazard detection; and ii) evaluating and comparing the framework's performance across different state-of-the-art VLMs, including Gemini, GPT-4o, Llama 3.2 Vision, and InternVL2.

## 3. Proposed Framework

In this paper, we propose and evaluate a comprehensive framework for construction hazard detection that integrates VLM with advanced prompt engineering techniques. As shown in Figure 2, the proposed framework consists of two interconnected modules: prompt engineering by analyzing safety regulations and VLM-based hazard inference. The methodology also includes a comprehensive evaluation of VLMs using quantitative metrics to assess the effectiveness and generalizability of the framework.

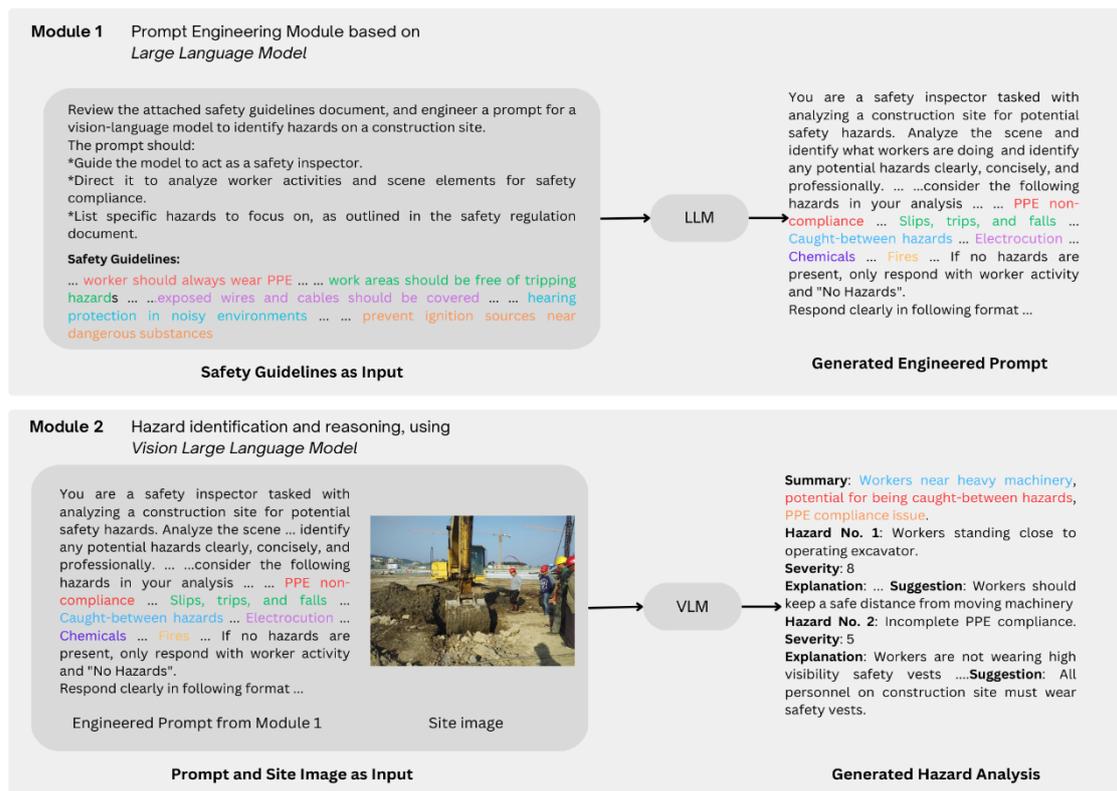

**Figure 2.** An overview of the framework's modules.

### 3.1. Prompt Engineering module

Prompt engineering is the process of crafting and refining inputs—known as prompts—to guide generative AI models in producing specific, high-quality outputs [48]. This involves designing instructions, questions, or cues that influence the AI's responses, ensuring they are relevant, accurate, and contextually appropriate. Studies have shown that VLMs exhibit improved performance when provided with well-engineered prompts [49,50]. Additionally, language models themselves can be leveraged to generate or refine prompts, allowing for a more dynamic and adaptive prompt engineering process[51,52].

Safety guidelines for construction sites contain detailed information about standards and protocols for mitigating risk and ensuring compliance. The Prompt Engineering Module serves as the first step of hazard inference, where safety guidelines text is parsed to identify actionable insights and structured into prompts that can effectively guide the VLM in the second module. It achieves this by passing the safety guidelines text to an LLM, along with a predefined prompt to engineer a prompt for the second module. This module ensures that the safety guidelines are correctly interpreted into prompts, guiding the VLM to assess construction site images for hazard detection.

*3.2. Hazard identification and reasoning module*

This module is responsible for analyzing construction site images and extracting hazard-related information. It receives the engineered prompt from the first module and inputs both the construction site image and the prompt into a VLM. The VLM analyzes the scene by detecting objects, evaluating spatial relationships, and interpreting contextual interactions. These findings are then mapped to the predefined safety criteria outlined in the prompt. Finally, the detected information is formatted into a structured response, providing a concise summary of hazards, their explanations, severity ratings, and actionable suggestions for mitigation. This proposed framework ensures a structured and scalable methodology for integrating VLMs into construction hazard detection, providing a practical foundation for automated safety monitoring in dynamic site conditions.

## 4. Experimental Validation

This section explains the experimental validation of the proposed framework, outlining the implementation details, dataset preparation, evaluation of models, and performance analysis.

*4.1. Framework implementation and validation*

To assess the feasibility of the proposed framework, we developed and deployed the framework as a prototype application. Figure 3 shows the dashboard of the developed application. The application was designed to engineer a prompt using safety guidelines, process construction site images, and generate structured hazard assessments aligned with safety guidelines using VLM-based reasoning. We used GPT-4-turbo as the backbone for this application. To guide VLM in hazard identification, we first engineered a structured prompt. We asked GPT-4-turbo to generate this prompt based on safety guidelines. These guidelines provided essential details about different hazard types and their conditions. Table 1 presents the safety guidelines we used in this study.

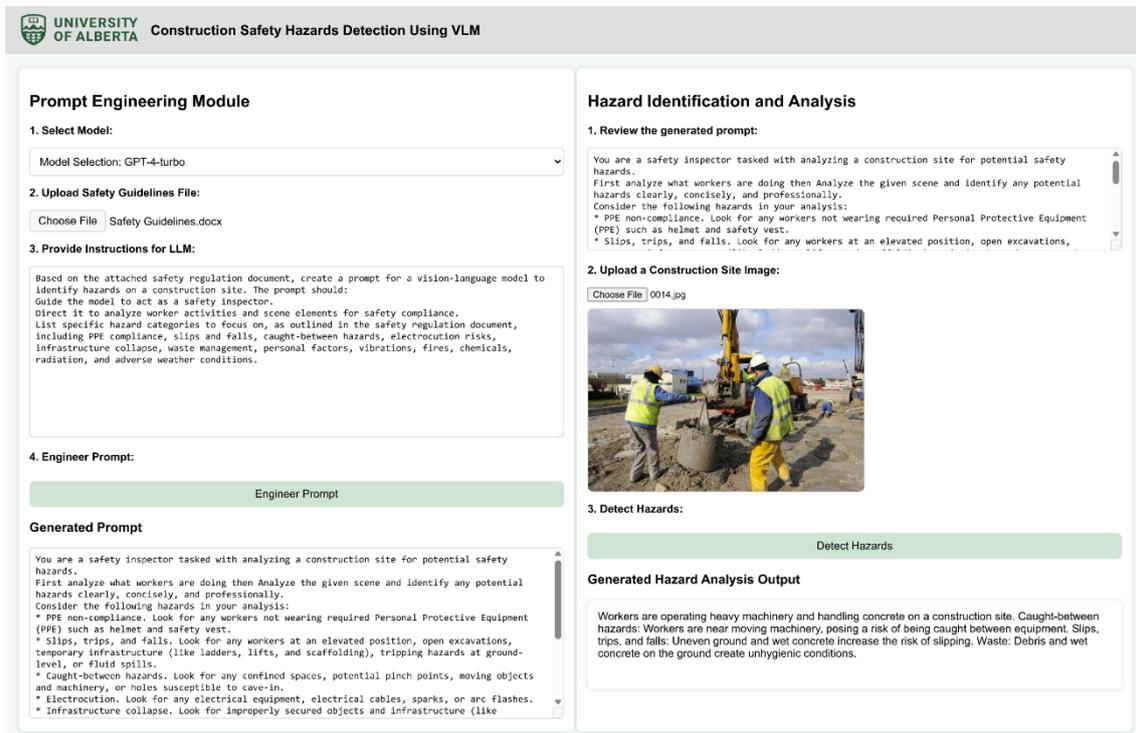

**Figure 3.** Dashboard of the developed application.

**Table 1.** Safety guidelines used in this study

| Sr. | Hazard | Conditions to look for |
| --- | --- | --- |
| 1 | PPE Non-Compliance | Workers not wearing required PPE (e.g., helmets, vests). |
| 2 | Slips, Trips, Falls | Workers at elevated positions, open excavations, temporary infrastructure (ladders, lifts, scaffolding), ground-level tripping hazards, or fluid spills. |
| 3 | Caught-Between Hazards | Confined spaces, potential pinch points, moving machinery or objects, or holes susceptible to cave-ins. |
| 4 | Electrocution | Electrical equipment, exposed electrical cables, arc flashes, etc. |
| 5 | Infrastructure Collapse | Improperly secured objects and infrastructure (e.g., scaffolding, ladders, excavations). |
| 6 | Waste | Unhygienic conditions, debris, dust, garbage needing disposal, etc. |
| 7 | Personal Factors | Unsafe work practices, horseplay, fatigued workers. |
| 8 | Vibrations | Sources of vibrations or excessive noise. |
| 9 | Fires | Active fires, fire-sparking sources, or flammable materials. |
| 10 | Adverse Weather Conditions | Snowy conditions, Thunderstorms, poor visibility. |

Once the prompt was generated, we paired it with the construction site image and passed it to the VLM model for analysis. The model successfully analyzed each image, identified hazards according to the guidelines leveraging its vision capabilities, and generated a detailed hazard description. We tested the framework under various site conditions, including worker activities, heavy machinery operations, and unsafe environments, to evaluate its adaptability across diverse scenarios. The responses were manually reviewed by the authors to verify their accuracy and consistency with safety guidelines. This step validated the initial feasibility of the framework in accurately identifying construction hazards. Figure 4 illustrates different construction site images and corresponding hazard assessments using the proposed framework. Following implementation and initial validation, we conducted a performance evaluation of multiple VLM models using a custom test dataset. The details of this evaluation are discussed in the subsequent sections.

| Image | GPT-4-turbo response |
|---|---|
| 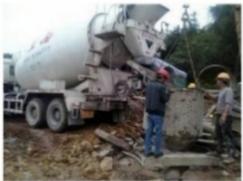 | Workers are operating near a concrete mixer truck and handling a suspended concrete bucket. One worker is not wearing a safety vest, indicating PPE non-compliance. The uneven ground and scattered debris present tripping hazards. The suspended concrete bucket poses a caught-between hazard. |
| 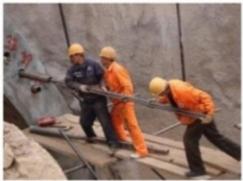 | Workers are maneuvering a large pipe on a makeshift wooden platform. The platform appears unstable, posing a risk of slips, trips, and falls. The workers are at risk of being caught between the pipe and the wall. |
| 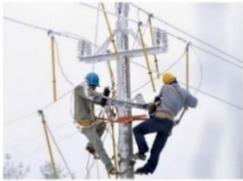 | Two workers are repairing electrical lines on a utility pole in snowy conditions. Elevated position poses a risk of slips, trips, and falls. Snowy weather may lead to poor visibility and slippery surfaces. Proximity to electrical lines presents a risk of electrocution. |
| 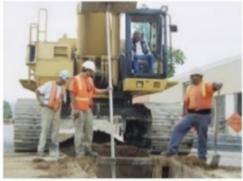 | Workers are measuring and inspecting an open trench with an excavator nearby. The open trench presents a risk of slips, trips, and falls. The trench also poses a caught-between hazard due to potential cave-ins. |
| 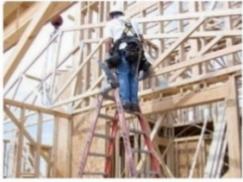 | The worker is on a ladder installing wooden frames at an elevated position. The ladder is not secured, posing a risk of slips, trips, and falls. The worker is wearing PPE, including a helmet and safety harness, so no PPE non-compliance is observed. No hazards are present. |
| 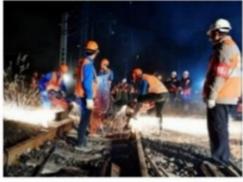 | Workers are engaged in cutting metal on railway tracks using power tools, producing sparks. Sparks from cutting operations present a fire hazard. |

**Figure 4.** Examples of hazard assessments generated by GPT-4-turbo

*4.2. Evaluation and Performance Comparison among VLMs*

To systematically evaluate the performance of VLMs in construction hazard detection, we developed a custom test dataset, implemented multiple VLMs, and assessed their accuracy and efficiency using quantitative metrics. This subsection explains the process of dataset preparation, model selection, implementing different VLMs, and evaluating their performance. Figure 5 illustrates the overall experimental pipeline for this study.

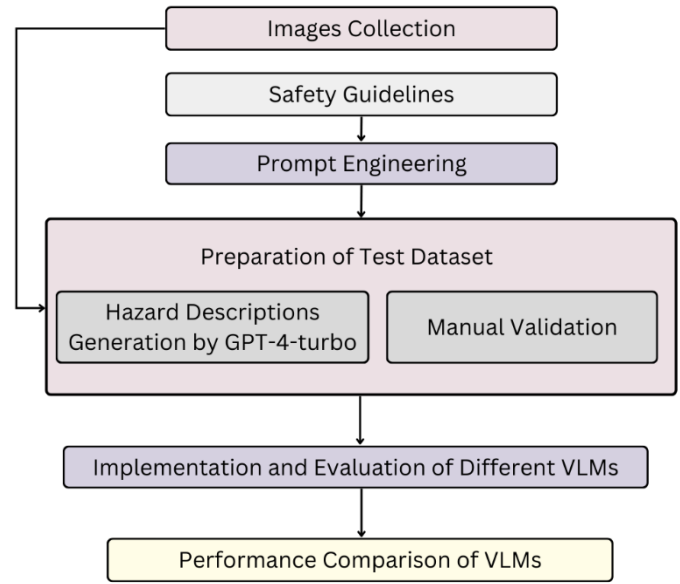

**Figure 5.** Overview of the performance comparison among VLMs

### 4.2.1 Experimental Dataset Creation

Existing benchmark datasets for VLMs typically consist of general images that do not adequately represent the complexities of construction site scenes. To address this gap, a custom dataset tailored to construction site scenarios was developed for this study. The dataset comprises a total of 1,100 images, each paired with detailed descriptions of hazards and corresponding suggestions for mitigation. In this study, we employed human-AI collaboration to generate the test dataset, a technique increasingly adopted across various domains. First, construction site images were sourced from publicly available datasets for construction safety hosted on Roboflow. These included the Construction Site Safety Dataset [53], Hard Hat Workers Dataset [54], and Safety Inspection Dataset [55]. These datasets provided a wide variety of images depicting workers, equipment, and safety scenarios relevant to construction sites. Once compiled, we filtered out images to exclude irrelevant or low-quality images, resulting in a curated set of 1100 good-quality images representing diverse site conditions.

Following image collection, a prompt was engineered using GPT-4-turbo, incorporating the same safety guidelines outlined in Table 2. A response output template was integrated into the engineered prompt to standardize the model's outputs. This template included fields for a list of identified hazards, followed by a severity rating, explanation, and actionable suggestion for each detected hazard. Next, the hazard descriptions for the images were generated using GPT-4-turbo. The engineered prompt designed for safety hazard analysis was input along with each image to produce a detailed description. To ensure consistency and relevance, the response generation process was carefully controlled, with output responses limited to a maximum of 250 tokens and a temperature setting of 0.3. This token limit was set to balance detail and relevance, ensuring that descriptions captured essential information without becoming overly verbose. A self-coded Python script was used to interface with the OpenAI GPT-4-turbo API, streamlining the text generation process for all images in the dataset. Figure 6 shows the process of dataset generation using GPT-4-turbo.

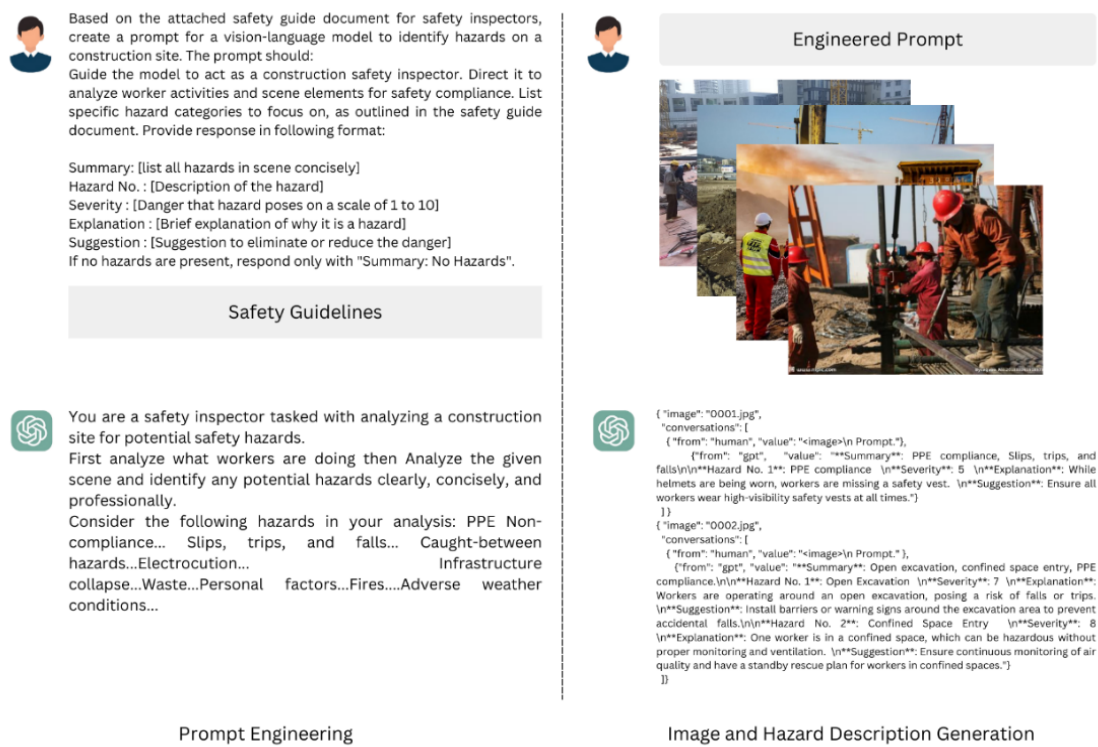

**Figure 6.** Dataset generation using GPT-4-turbo. (Left) Prompt engineering. (Right) Image captions with hazard analysis generation using the engineered prompt.

Once generated, the responses were manually reviewed by a human expert with knowledge of construction safety guidelines. Each image and its corresponding hazard description were carefully examined to verify the correctness, relevance, and clarity of the model-generated outputs. Misclassifications, missing hazards, or vague descriptions were identified and refined to align with established safety guidelines. In cases where the model's output failed to capture critical hazards or misrepresented safety risks, adjustments were made to ensure accuracy and completeness. This validation step ensured that all identified hazards, severity ratings, and suggestions were accurate and contextually appropriate. This manual review step was crucial for refining the dataset and ensuring that it met the high standards required for effective hazard detection and mitigation.

4.2.2. Model implementation and evaluation

In this section, we implement and evaluate state-of-the-art VLMs for construction hazard detection. All models were tasked with performing hazard analysis using the same engineered prompt and the curated test dataset images. We used Gemini 2 flash, 1.5 Pro and 1.5 flash-8B, GPT-4o and mini, InternVL2 1B, 2B, 4B and 8B, and Llama 3.2 Vision 8B in this study. The rationale behind selecting these models is their benchmark performance and to introduce sufficient diversity, ensuring the identification of the most effective model for construction hazard analysis. This diversity ensures that the study covers a range of deployment scenarios, from lightweight open-source models to high-performance cloud-based proprietary models.

Intern VL2 and Llama 3.2 Vision are open-sourced, which provides the flexibility to customize the models as needed and host them in-house for better control and efficiency. This eliminates the privacy and regulatory risks associated with sending data to third-party APIs. Moreover, in-house hosting reduces the operation and opportunity costs factor, as commercial APIs often incur significant expenses and impose strict rates and token limits. By hosting the models locally, the cost becomes relatively fixed compared to the variable costs associated with API usage. Among the open-source VLMs, smaller models in the 1B-4B parameter range were included, making them suitable for resource-constrained environments and on-device AI deployment. While large models with more parameters typically deliver higher performance, they are also computationally resource intensive. On the other hand, proprietary models like Gemini-1.5 Pro and GPT-4o demonstrate superior accuracy and performance, making them ideal for applications that require state-of-the-art capabilities without concerns about hosting or API costs.

We implemented and evaluated the open-source models in a Python Environment using libraries such as HuggingFace Transformers [56], adhering to the guidelines provided in their respective model cards. These models were implemented and tested on a high-performance workstation equipped with an Intel Xeon Gold 6242R processor, 128 GB RAM, and an NVIDIA A6000 GPU. For GPT-4o and Mini, we used the OpenAI API [57], and for the Gemini models, the Gemini API [58] was employed. Each model was deployed using a consistent experimental setup, with a temperature set to 0.3 and a maximum token limit of 250 to ensure uniform response characteristics. The test dataset images, and engineered prompts were used as input, and each model generated hazard assessments in a predefined structured format for analysis. The generated outputs were then compared against the ground truth dataset created earlier in this study. This uniform approach allowed for a direct and objective comparison of their performance in the subsequent evaluation phase.

4.2.3. Evaluation Metrics

To evaluate the performance of our construction safety hazard detection framework, two primary evaluation objectives were defined: (1) accuracy of the hazard detection, and (2) Overall response accuracy and completeness. In this study, the evaluation was conducted using quantitative metrics commonly used for Language models, as detailed below.

4.2.3.1. Cosine Similarity

Cosine similarity is a common metric used to evaluate the overall contextual understanding of the model. It measures the semantic closeness between two textual responses by comparing their vectorized embeddings. For this study, we computed cosine similarity using Sentence-BERT embeddings, specifically leveraging the 'paraphrase-MiniLM-L12-v2' model. The formula for cosine similarity is given by:

$$Cosine\ Similarity = \frac{\vec{A}.\vec{B}}{\|\vec{A}\|.\|\vec{B}\|} \tag{1}$$

Here, A and B are embedding vectors of the model-generated response and the ground truth response, respectively. Cosine similarity ranges from -1 (completely dissimilar) to 1 (perfect alignment). A higher cosine similarity score indicates that the model-generated hazard descriptions are semantically closer to the ground truth annotations. that ensure robust semantic evaluation.

4.2.3.2. BERTScore

BERTScore evaluates the semantic similarity of two textual responses at the word embedding level. Unlike traditional n-gram metrics, BERTScore leverages the contextual embeddings generated by BERT (Bidirectional Encoder Representations from Transformers) to compute precision, recall, and F1 scores for textual comparisons. For this study, we computed BERTScore using the pre-trained 'Roberta-large' model. The BERTScore precision, recall, and F1 are calculated as:

$$Precision = \frac{\sum_{t \in T_p} max_{g \in T_p} sim(t, g)}{|T_p|} \tag{2}$$

$$Recall = \frac{\sum_{g \in T_g} max_{g \in T_p} sim(g, t)}{|T_g|} \tag{3}$$

$$BERTScoreF1 = \frac{2.Precision.Recall}{Precision + Recall} \tag{4}$$

where $T_p$ and $T_g$ represent the set of tokens in predicted response, and ground truth respectively and sim(t,g) is the similarity between token t and g based on their embeddings.

4.2.3.3. Judge Model Evaluation

In addition to computational metrics, we also used LLM-as-a-Judge evaluation to compare the model-generated outputs against ground truth, a technique increasingly adopted in recent studies [59–61] evaluating generated responses across multiple qualitative dimensions. This method provides an additional layer of validation beyond numerical similarity scores, that offers insights into how well models capture context, reasoning, and completeness in hazard descriptions.

For this study, we used Gemini-1.5 Pro as a judge model [62] to systematically score the responses against ground truth based on predefined criteria. The model evaluated each response without any prior knowledge of its source, ensuring an objective and unbiased assessment process. The evaluation focused on three key dimensions:

1. *Completeness:* Ensures that all relevant hazards present in the ground truth are identified in the predicted response.
2. *Accuracy:* Measures the correctness and relevance of the identified hazards, severity ratings, and mitigation suggestions.
3. *Clarity*: Evaluates the professionalism, conciseness, and overall readability of the model's output.

The Gemini model was instructed using a structured prompt to assign a score for each criterion. Each criterion was scored on a scale of 1 to 5, and the final evaluation was formalized using a Normalized score as follows:

$$Normalized\ Score = \frac{Completeness + Accuracy + Clarity}{Number\ of\ criteron\ \times\ Highest\ score} \quad (5)$$

## 5. Results

Examples of hazard detection using different VLM models are shown in Figure 7. The performance of each model was assessed using the metrics discussed earlier. The evaluation results for models are presented in Table 2. Additionally, the mode of deployment and recorded average inference speed for each model are summarized in Table 3.

| Image | Predicted response | Ground Truth | Evaluation |
|---|---|---|---|
| 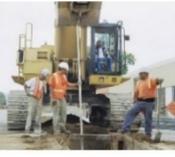 | [GPT-4o]<br>Summary: Open excavation, proximity to moving machinery.<br>Hazard No. 1: Open excavation Severity: 8 Explanation: Workers are near an open trench, creating a risk of falling or being caught in a cave-in. Suggestion: Install barriers and warning signs around the trench to prevent falls. Ensure proper trench shoring.<br>Hazard No. 2: Proximity to moving machinery Severity: 7 Explanation: Workers are close to heavy machinery, which poses a risk of being struck or caught between moving parts. Suggestion: Ensure clear communication and use of spotters to maintain safe distances. Consider machine guards if possible. | Summary: Open trench, potential caught-between hazard<br>Hazard No. 1: Open trench. Severity: 7. Explanation: The open trench presents a risk of falls or being caught in a collapse. Workers near the edge without proper fall protection are at risk. Suggestion: Install barriers and warning signs around the trench and ensure proper training and fall protection for workers near the edge.<br>Hazard No. 2: Potential caught-between hazard (machinery). Severity: 6. Explanation: Workers are in close proximity to heavy machinery, which could result in caught-between incidents if the machinery moves unexpectedly. Suggestion: Establish clear communication protocols and exclusion zones around operating machinery. | Cosine Similarity: 0.84<br>BERTScore F1: 0.94<br>LLM as Judge: 0.88 |
| 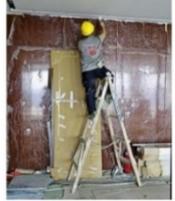 | [Gemini-1.5-Pro]<br>Summary: Unstable ladder positioning, potential for slips, trips, and falls.<br>Hazard No. 1: Unstable ladder positioning Severity: 8 Explanation: The ladder is not positioned securely on a flat surface, and the worker is standing near the top, increasing the risk of falling. Suggestion: Ensure the ladder is placed on a stable, flat surface and encourage the use of an extension ladder or scaffolding for better stability.<br>Hazard No. 2: Clutter and debris on the ground Severity: 6 Explanation: Debris on the ground can cause trips and falls, leading to injuries. Suggestion: Clear the area of debris and clutter to ensure a safe working environment. | Summary: Worker at height using an unsecured ladder; debris and material waste scattered across the floor; lack of eye protection.<br>Hazard No. 1: Unsecured ladder being used at height Severity: 7 Explanation: The ladder is not secured and could shift or fall, leading to a fall from height. Suggestion: Secure the ladder with appropriate restraints or have a spotter hold the base.<br>Hazard No. 2: Debris and material waste scattered on the floor Severity: 5 Explanation: Trip hazards present from scattered materials on the floor. Suggestion: Clear the floor of debris and waste materials regularly. Implement a clean-as-you-go policy. | Cosine Similarity: 0.77<br>BERTScore F1: 0.89<br>LLM as Judge: 0.86 |
| 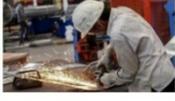 | [Llama-3.2-11B-Vision]<br>Summary: Potential hazards include lack of PPE, sparks, electrical equipment, and unhygienic conditions.<br>Hazard No. 1: PPE Compliance Severity: 8/10 Explanation: Worker is not wearing safety glasses, posing a significant risk of head injuries from sparks. Suggestion: Ensure the worker wears appropriate PPE, including a helmet and safety glasses.<br>Hazard No. 2: Sparks Severity: 9/10 Explanation: Sparks from the grinding tool pose a risk of fire or electrocution if they come into contact with flammable materials or exposed wiring. Suggestion: Use a spark arrestor or take measures to contain sparks to prevent fires or electrical hazards.<br>Hazard No. 3: Electrical Equipment Severity: 7/10 Explanation: Operating a power tool near electrical equipment increases the risk of electrical shock if there is a fault or short circuit. Suggestion: Ensure that the electrical equipment is properly grounded and maintained to prevent electrical hazards. | Summary: PPE non-compliance (missing eye protection), sparks (fire hazard), potential noise hazard.<br>Hazard No. 1: PPE non-compliance (missing eye protection) Severity: 7 Explanation: The worker is not wearing eye protection while operating machinery that generates sparks, posing a risk of eye injury. Suggestion: Ensure all workers wear appropriate eye protection such as safety goggles when operating machinery involving sparks.<br>Hazard No. 2: Sparks (fire hazard) Severity: 6 Explanation: The sparks emitted during metal grinding can ignite flammable materials. Suggestion: Clear the work area of any flammable materials and have fire extinguishers readily available.<br>Hazard No. 3: Potential noise hazard Severity: 5 Explanation: Grinding operations can produce excessive noise, potentially damaging hearing over time. Suggestion: Provide hearing protection like earplugs or earmuffs to workers operating noisy machinery. | Cosine Similarity: 0.82<br>BERTScore F1: 0.87<br>LLM as Judge: 0.67 |
| 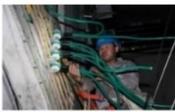 | [InternVL2-8B]<br>Summary: Exposed electrical cables.<br>Hazard 1: Electrocution. Severity: 8. Explanation: Exposed electrical cables pose a significant risk of shock or electrocution, especially if workers handle them without proper insulating tools or equipment. Suggestion: Ensure all workers are equipped with personal insulating mats and use insulated tools when working with exposed electrical cables. | Summary: Electrocution, PPE compliance<br>Hazard No. 1: Electrocution. Severity: 9. Explanation: The worker is handling electrical cables, which poses a high risk of electrocution if they are live and not properly insulated. Suggestion: Ensure that power is turned off before maintenance, use insulated tools, and verify cable insulation integrity.<br>Hazard No. 2: PPE compliance. Severity: 6. Explanation: The worker is wearing a helmet but may need additional PPE such as gloves, face shields, or insulated clothing when working with electrical cables. Suggestion: Provide and enforce the use of appropriate PPE for electrical work, such as rubber gloves and face protection. | Cosine Similarity: 0.90<br>BERTScore F1: 0.91<br>LLM as Judge: 0.72 |

**Figure 7.** Examples of models' response compared with ground truth and evaluation

**Table 2.** Evaluation Results for different VLMs.

| Models | Hazard Detection Accuracy | | Overall Response Accuracy and Completeness | | |
|---|---|---|---|---|---|
| | Cosine Similarity | BERTScore F1 | Cosine Similarity | BERTScore F1 | LLM as Judge |
| Gemini 2 flash | 0.537 | 0.869 | 0.674 | 0.888 | 0.612 |
| Gemini 1.5 Pro | 0.550 | 0.875 | 0.729 | 0.888 | 0.671 |
| Gemini 1.5 Flash 8B | 0.492 | 0.874 | 0.711 | 0.891 | 0.634 |
| GPT-4o | 0.551 | 0.877 | 0.730 | 0.906 | 0.612 |
| GPT-4o mini | 0.532 | 0.866 | 0.700 | 0.892 | 0.523 |
| Llama-3.2-11B-Vision | 0.504 | 0.850 | 0.684 | 0.859 | 0.480 |
| Intern VL2 8B | 0.501 | 0.860 | 0.700 | 0.881 | 0.531 |
| Intern VL2 4B | 0.465 | 0.855 | 0.667 | 0.872 | 0.491 |
| Intern VL2 2B | 0.358 | 0.795 | 0.658 | 0.866 | 0.439 |
| Intern VL2 1B | 0.381 | 0.766 | 0.622 | 0.857 | 0.361 |

*5.1. Hazard Detection Accuracy*

Hazard detection accuracy measures the ability of each VLM model to correctly identify hazards present in construction scene images, comparing model-generated output to ground truth descriptions. Figure 8 shows the accuracy of hazard detection across models.

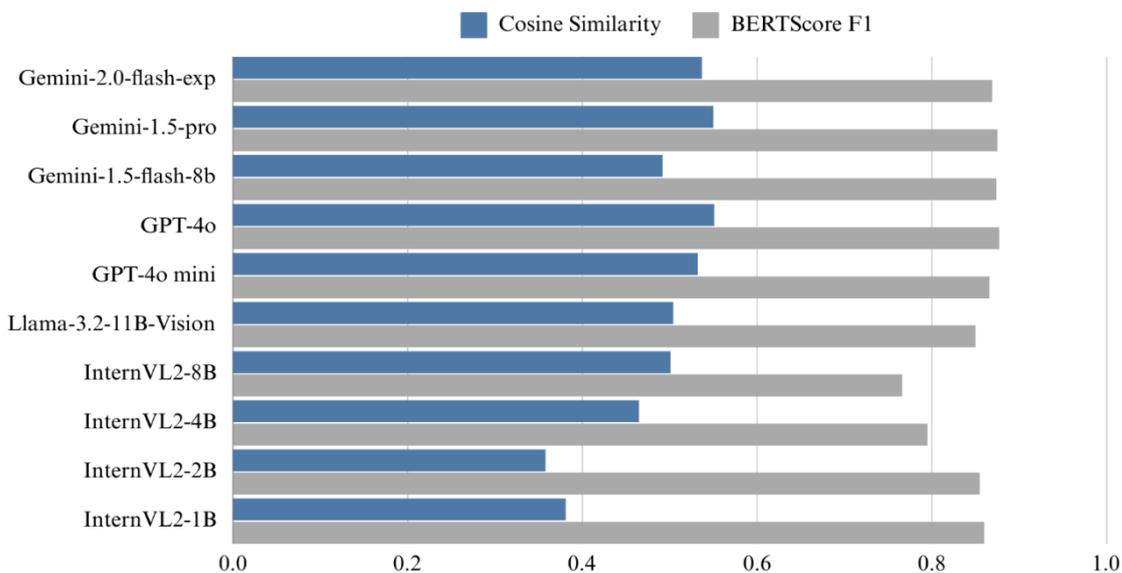

**Figure 8.** Evaluation results for accuracy of hazard detection

GPT-4o achieved the highest accuracy among all models, with a Cosine Similarity of 0.551 and BERTScoreF1 of 0.877. Gemini 1.5 Pro delivered highly competitive results, with a Cosine Similarity of 0.550 and BERTScoreF1 of 0.860. These results highlight the superior capabilities of these models in understanding construction site hazards and generating accurate responses. The open-source models, particularly, Llama-3.2-11B-Vision and InternVL2-8B, also showed comparable performances. Llama-3.2-11B-Vision achieved a Cosine Similarity of 0.504 and a BERTScoreF1 of 0.850, while InternVL2-8B had a Cosine Similarity of 0.501 and 0.860. Smaller models such as InternVL2 1B and 2B, performed relatively poorly with Cosine Similarity scores of 0.381 and 0.358, respectively. This performance gap can be attributed to their reduced capacity to understand complex spatial relationships, as smaller parameter sizes limit the model's ability to generalize well across diverse scenarios.

*5.2. Overall Response Accuracy and Completeness*

To assess overall response accuracy and completeness, we used LLM-as-a-Judge evaluation in addition to Cosine Similarity and BERTScore. In this evaluation, we compared the models' understanding capabilities to declare an entity as a hazard. This not only measures the accuracy but also the reasoning and contextual relevance of the entire response generated by the model.

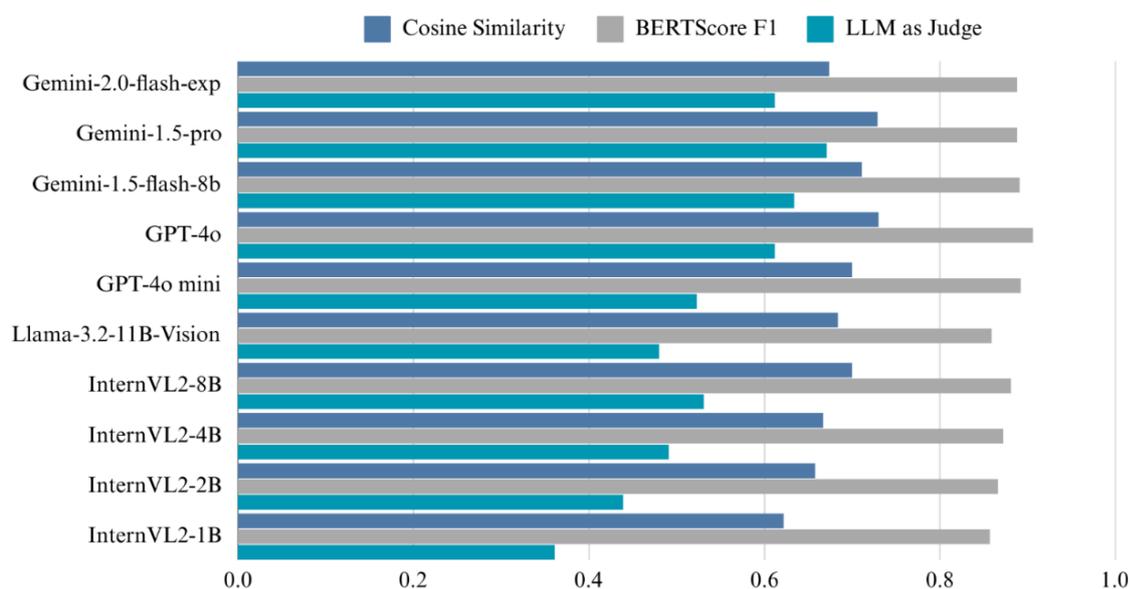

**Figure 9.** Evaluation results for overall response accuracy and completeness

As shown in Fig 9, GPT-4o achieved the highest performance among all models, with an overall Cosine Similarity of 0.730 and BERTScoreF1 of 0.906, followed closely by Gemini 1.5 Pro, with 0.729 Cosine Similarity and BERTScore F1 of 0.888. This highlights that these models are capable to generate accurate and contextually rich responses that align closely with ground truth descriptions for hazards. Mid-tier models including Gemini 1.5 flash 8B, GPT-4o-mini, and Gemini 2 flash-exp, provided a balance between performance, cost, and efficiency, with Cosine Similarity scores of 0.711, 0.700, and 0.674 respectively. Among open-source models, InternVL2-8B delivered the strongest performance, achieving a Cosine Similarity value of 0.700 and BERTScoreF1 of 0.881. Llama-3.2-8B-Vision also demonstrated reliable accuracy, with a Cosine Similarity of 0.684 and a BERTScore F1 of 0.859.

In LLM-as-Judge evaluation, Gemini models excelled in assessing completeness and reasoning, with Gemini 1.5 Pro securing the highest score of 0.671, indicating strong contextual understanding and response clarity. GPT-4o achieved a score of 0.612, further validating its robustness in generating comprehensive hazard assessments. In open-source models, InternVL2-8B reached an impressive score of 0.531, surpassing the mid-tier GPT model.

In conclusion, leveraging vision Language Models for automatic safety hazard identification shows significant promise, as evidenced by the diverse evaluation metrics and performance outcomes discussed. Proprietary models such as GPT-4o and Gemini 1.5 Pro demonstrated superior contextual understanding and accuracy, and open-source models like InternVL2-8B and Llama 3.2-8B provided competitive and cost-efficient alternatives, particularly for local and resource-constrained deployments. However, this approach has its limitations and challenges. We will discuss the details of failure cases and limitations in the subsequent section, emphasizing the need for further enhancement in research and refinement to improve both the accuracy and efficiency of the framework.

*5.3. Inference Speed and Computational Efficiency*

Inference speed is a crucial factor in evaluating the practical deployment feasibility of VLMs for real-time hazard detection. The time required to process a single image varies significantly across different VLMs due to differences in their model architecture, parameter size, and deployment mode. Table 3 summarizes the average inference speed for models.

**Table 3.** Mode of deployment and recorded average inference speed for analyzing one image.

| Model | Deployment Mode | Inference Speed (seconds) |
|---|---|---|
| Gemini 2 flash | Google Vertex AI Cloud API | 0.94 |
| Gemini 1.5 Pro | Google Vertex AI Cloud API | 2.94 |
| Gemini 1.5 Flash 8B | Google Vertex AI Cloud API | 0.86 |
| GPT-4o | OpenAI Cloud API | 4.57 |
| GPT-4o mini | OpenAI Cloud API | 3.18 |
| Llama-3.2-11B-Vision | Locally hosted on NVIDIA A6000 | 8.40 |
| Intern VL2 8B | Locally hosted on NVIDIA A6000 | 5.30 |
| Intern VL2 4B | Locally hosted on NVIDIA A6000 | 4.80 |
| Intern VL2 2B | Locally hosted on NVIDIA A6000 | 3.41 |
| Intern VL2 1B | Locally hosted on NVIDIA A6000 | 3.24 |

GPT-4o exhibited the highest latency (4.57 seconds per image), whereas Gemini 2 Flash delivered the fastest response (0.94 seconds per image), making it more suitable for real-time applications. Similarly, Gemini 1.5 Flash 8B achieved an inference time of 0.86 seconds, highlighting the advantages of smaller, optimized models. For open-source models hosted locally, inference speed was influenced by both model size and hardware resources. Larger models, such as Llama-3.2-11B-Vision and InternVL2-8B, exhibited significantly higher latency, with Llama-3.2-11B-Vision taking 8.4 seconds per image and InternVL2-8B taking 5.30 seconds when deployed on an NVIDIA A6000 GPU. Smaller models, such as InternVL2-1B and InternVL2-2B, demonstrated faster inference speeds of 3.24 seconds and 3.41 seconds, respectively, making them more suitable for edge computing and resource-constrained environments. The trade-off between inference speed and accuracy is evident in the results. While larger models provide superior hazard detection accuracy, they are computationally intensive and may not be ideal for real-time safety monitoring applications. In contrast, smaller models and optimized cloud-based solutions offer significantly lower latency, making them preferable for on-site deployment scenarios where near-instantaneous hazard detection is required.

*5.4. Failure Cases*

Despite the promising results, the framework encountered several failure cases during evaluation. We identified three major causes for the failure cases: (1) false identification of hazards; (2) difficulty in identifying context-specific hazards; and (3) model hallucination. False responses generated due to these errors are presented in Table 4.

**Table 4.** Examples of Failure Cases in Hazard Detection (false response is underlined)

| Failure Causes | Image | Identified Hazards |
|---|---|---|
| False Hazards | 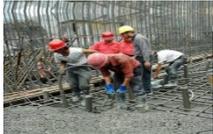 | [GPT-4o] <u>One worker is not wearing a safety helmet.</u> Workers are standing on wet concrete, risking slips and falls. Open rebar poses a trip hazard. <u>Rebar structure appears unsecured, posing a risk of collapse.</u> |
| False Hazard, Context-Specific Hazard Misclassification | 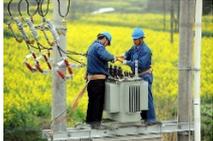 | [Gemini 1.5 Pro] Electrical hazards, working at height, <u>lack of fall protection</u> |
| Model Hallucination | 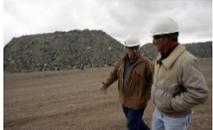 | [GPT-4o] <u>Workers at heights without proper fall protection</u>, improper PPE use, debris posing trip hazards, <u>exposed electrical cables</u>, and <u>poorly secured scaffolding.</u> |
| Model Hallucination | 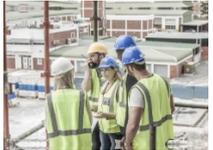 | [Gemini 1.5 Pro] <u>Scaffolding inadequately secured</u>, <u>fall hazards from height</u>, <u>debris/waste</u> |
| False Hazard, Model Hallucination | 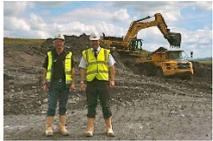 | [Llama 3.2-8B] <u>Unsecured excavator. Inadequate PPE (no hard hats).</u> <u>Unsecured and unmarked open excavations.</u> <u>Unhygienic conditions.</u> <u>Unsecured and unmarked ladders. Unsecured and unmarked scaffolding.</u> |

5.4.1. False identification of hazards

This error occurs when the system misclassifies safe conditions as hazardous, leading to false positives. For example, properly anchored scaffolding or well-secured machinery was sometimes flagged as unstable infrastructure. Such errors undermine the accuracy and contextual understanding of the models.

5.4.2. Difficulty in identifying context-specific hazards

The framework struggled with nuanced and context-depended hazards. This error is particularly evident for 'workers working at height' scenarios. For example, the models often failed to distinguish between missing safety harnesses and correctly secured ones. Similarly, in cases where workers appeared to be standing on scaffolding, the models might be unable to determine whether adequate fall protection measures were in place, highlighting their limited ability to interpret fine-grained spatial relationships or usage contexts.

5.4.3. Model hallucination

Hallucination is a known limitation of vision language models, generating responses about objects not present in the given images [63]. This issue was observed during the evaluation of the models for hazard detection, exhibiting a tendency to overgeneralize and introduce irrelevant information into response. For instance, as shown in Table 4, the models occasionally flagged non-existent elements in the image as hazards.

## 6. Discussion

### 6.1. Impact and Feasibility

To improve the accuracy and efficiency of construction safety management systems, this study proposes the use of vision language models for hazard identification. This integration offers a transformative and practical approach to address challenges in traditional safety monitoring systems, providing a contextual understanding of dynamic construction environments. Results from this study validate the feasibility of applying VLMs for construction safety monitoring. Commercial models such as GPT-4o and Gemini 1.5 Pro demonstrate exceptional performance in hazard identification and contextual understanding. Meanwhile, open-source alternatives such as InternVL2 and Llama 3.2 Vision offer scalable solutions with adaptable deployment options.

The feasibility of real-time deployment depends on computational efficiency and inference speed. While cloud-based models offer faster processing time, their reliance on network stability and privacy concerns [64] may limit their suitability for sensitive or remote applications. Locally hosted models address these challenges but require significant GPU resources. Further enhancements should focus on resource-optimized VLMs and other efficiency-improving techniques, such as parallel processing, model quantization, and edge computing integration, to achieve a balance between real-time performance and detection accuracy.

*6.2. Methodological advances*

This study represents a significant advancement in the methodological integration of vision and language capabilities for construction hazard identification. Unlike traditional entity detection techniques, VLMs employ semantic reasoning to analyze spatial relationships and contextual dependencies between site elements. By utilizing prompt engineering techniques, this framework can extract relevant information and generate actionable insights, including hazard severity assessments and mitigation strategies. Moreover, the zero-shot capabilities of VLMs eliminate the need for extensive task-specific training, enabling the system to adapt to unseen hazards or new safety conditions.

*6.3. Practical applications and advantages*

The proposed VLM framework has immediate practical applications in construction safety monitoring. It provides an automated solution to manual safety monitoring, which is both inefficient and costly. The framework provides an adaptable method, which allows the users to customize input prompts based on specific safety guidelines, ensuring compliance with varying regulatory requirements. The framework enables proactive responses to dynamic risks, enhancing overall site safety and regulatory compliance. Moreover, the framework can be seamlessly integrated into existing monitoring systems leveraging local resources or cloud computing, making it adaptable for projects of varying scale and complexity.

*6.4. Errors and Inaccuracies*

While the proposed method demonstrates substantial promise, it is not immune to errors and inaccuracies, particularly model hallucination errors. Furthermore, the models exhibit difficulty in interpreting fine-grained spatial relationships and operational contexts. Such inaccuracies can lead to misclassification of hazards, impacting the reliability of the system in certain scenarios. It is crucial to address these errors, perhaps through improved model architectures or training the models on diverse and representative construction site datasets.

*6.5. Limitations of the study*

This study proposes and evaluates the use of VLMs for the identification of general and context-specific hazards on construction sites. The limitations of this study are explained as follows:

(1) Although our study confirms the feasibility of the method, we need to admit the challenges identified in the evaluation should be addressed before practical implementation on construction sites. These include mitigating inaccuracies and model hallucinations by training on diverse construction site image datasets. Additionally, real-time performance constraints necessitate further optimization to ensure reliable deployment in dynamic environments.

(2) This study evaluated only a specific set of VLM models identified based on their benchmark performance. Furthermore, only open-source models with less than 20B parameters were considered due to limited GPU capacity. The exploration of other high-capacity models or fine-tuning using construction-specific datasets was beyond the scope of this research study.

These limitations highlight areas for future research focused on enhancing the effectiveness, accuracy, and applicability on construction sites.

## 7. Conclusion

This study introduced a framework leveraging VLMs to address the complex challenge of hazard identification with contextual understanding in construction safety management. By integrating the construction safety guidelines with visual data from sites, the proposed framework bridges the gap between traditional safety monitoring systems and modern AI capabilities. In the method, we passed the safety regulations for a construction site to a VLM to engineer a prompt. The prompt was then paired with site visuals to infer hazards and provide actionable suggestions using VLM. The proposed approach not only automates hazard detection but also addresses the key limitations in existing vision-based methods.

The proposed approach was validated by developing a prototype web application and experimental testing for different construction site scenarios. Following validation, different pre-trained VLMs were evaluated and compared for their accuracy in detecting construction hazards. GPT-4o and Gemini 1.5 Pro achieved exceptional performance in hazard identification and contextual understanding with Cosine Similarity scores of 0.730 and 0.739, respectively. Concurrently, open-source alternatives such as InternVL2 and Llama 3.2 Vision emerged as cost-efficient options, ensuring accessibility for resource-constrained environments. This dual-track approach ensures the adaptability of proposed solutions for different project scales and resource availabilities. From a real-time feasibility perspective, inference speed remains a critical bottleneck for deploying VLMs for real-time monitoring applications.

The contributions of this study are multifaceted. First, it underscores the potential of combining multimodal inputs to improve safety monitoring outcomes with greater contextual depth and accuracy. Second, it establishes the feasibility of using VLMs for proactive hazard identification. Third, the proposed method aligns with the growing need for automation in the construction industry and the rapid advancement of generative AI technologies, offering scalable and practicable solutions for integrating cutting-edge AI systems into real-world construction safety management practices. Looking forward, this research lays the groundwork for significant future research endeavors. Model inaccuracies, such as false positives and hallucination errors, highlight the need for further refinement in contextual reasoning and data alignment. Additionally, optimizing the trade-off between processing speed and detection accuracy remains a critical area for exploration. With ongoing advancements in multimodal large language models and the development of data centers for cloud computing technologies, the potential for scalable, high-performance hazard detection systems in construction safety management is becoming increasingly feasible. Future directions include fine-tuning models on construction site datasets, integrating domain-specific knowledge for improved contextual understanding and optimizing real-time processing capabilities. Furthermore, exploring hybrid Human-in-the-loop [65] approaches can enhance system reliability, while leveraging expert feedback to continuously improve model performance. These advancements will ensure the continued evolution of VLM-based systems, paving the way for safer, smarter, and more efficient construction practices.

**Declaration of Competing Interest**

The authors declare that they have no known competing financial interests or personal relationships that could have appeared to influence the work reported in this study.

**Acknowledgment**

This work was supported by the NSERC Alliance under Grant ALLRP 576826 – 22. The authors would also like to thank Mehmood Ahmad and Issac Joffee for their valuable insights and contributions to this research.

**Data availability**

Data will be made available on request.